\documentclass[a4paper]{article}
\pdfoutput=1
\usepackage{graphicx,times,amsmath}
\usepackage[noend]{algorithmic}
\usepackage[boxed]{algorithm}

\title{Multi-Dimensional Recurrent Neural Networks}

\author{Alex Graves, Santiago Fern\'andez, J\"urgen Schmidhuber \\
IDSIA\\
Galleria 2, 6928 Manno, Switzerland\\
\{alex,santiago,juergen\}@idsia.ch\\}

\begin{document}
\maketitle

\begin{abstract}
Recurrent neural networks (RNNs) have proved effective at one dimensional sequence learning tasks, such as speech and online handwriting recognition. Some of the properties that make RNNs suitable for such tasks, for example robustness to input warping, and the ability to access contextual information, are also desirable in multidimensional domains. However, there has so far been no direct way of applying RNNs to data with more than one spatio-temporal dimension. This paper introduces multi-dimensional recurrent neural networks (MDRNNs), thereby extending the potential applicability of RNNs to vision, video processing, medical imaging and many other areas, while avoiding the scaling problems that have plagued other multi-dimensional models. Experimental results are provided for two image segmentation tasks.
\end{abstract}

\section{Introduction}

Recurrent neural networks (RNNs) were originally developed as a way of extending neural networks to sequential data. Because of their recurrent connections, RNNs are able to make use of previous context. Moreover, because RNNs can adapt to stretched or compressed input patterns by varying the rate of change of their internal state, they are more robust to temporal warping than non-recursive models.

In recent experiments, RNNs have outperformed hidden Markov Models (HMMs) in a variety of speech and online handwriting recognition tasks \cite{graves05nn,graves05icann,graves06icml,liwicki07icdar}.

Access to contextual information and robustness to warping are also important when dealing with multi-dimensional data. For example, a face recognition algorithm should be able to access the entire face at once, and it should be robust to changes in perspective, distance etc. It therefore seems desirable to apply RNNs to such tasks. 

However, the RNN architectures used so far have been explicitly one dimensional, meaning that in order to use them for multi-dimensional tasks, the data must be pre-processed to one dimension, for example by presenting one vertical line of an image at a time to the network. 
The most successful use of neural networks for multi-dimensional data has been the application of convolution networks to image processing tasks such as digit recognition \cite{lecun-98,simard03icdar}. One disadvantage of convolution nets is that because they are not recurrent, they rely on hand specified kernel sizes to introduce context. Another disadvantage is that they don't scale well to large images. For example, sequences of handwritten digits must be pre-segmented into individual characters before they can be recognised by convolution nets \cite{lecun-98}.

Various statistical models have been proposed for multi-dimensional data, notably multi-dimensional HMMs. However, multi-dimensional HMMs suffer from two severe drawbacks: (1) the time required to run the Viterbi algorithm, and thereby calculate the optimal state sequences, grows exponentially with the number of data points; (2) the number of transition probabilities, and hence the required memory, grows exponentially with the data dimensionality. Numerous approximate methods have been proposed to alleviate one or both of these problems, including pseudo 2D and 3D HMMs \cite{hulsken01p3dhmm}, isolating elements \cite{univ00image}, approximate Viterbi algorithms \cite{joshi05mdhmm}, and dependency tree HMMs \cite{jiten06dthmm}. However, none of these methods are able to exploit the full multi-dimensional structure of the data. 

As we will see, multi dimensional recurrent neural networks (MDRNNs) bring the benefits of RNNs to multi-dimensional data, without suffering from the scaling problems described above.


Section \ref{sec:mdrnn} describes the MDRNN architecture, Section \ref{sec:exps} presents two experiments on image segmentation, and concluding remarks are given in Section \ref{sec:conclusion}.

\section{Multi-Dimensional Recurrent Neural Networks}\label{sec:mdrnn}

\begin{figure}
  \begin{minipage}[t]{.45\textwidth}
    \begin{center}  
      \includegraphics[width=\columnwidth]{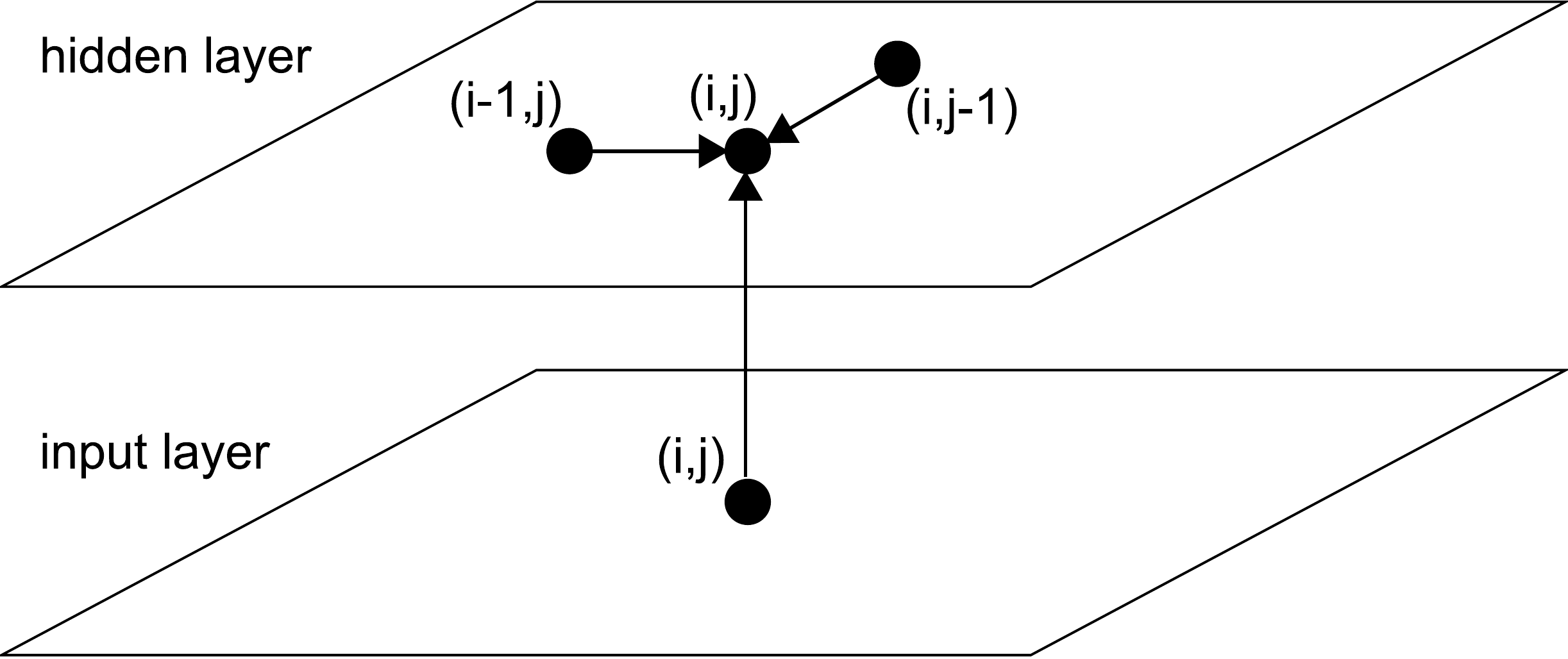}
      \caption{2D RNN Forward pass.} 
      \label{fig:mdrnn_forward}
    \end{center}
  \end{minipage}
  \hfill
  \begin{minipage}[t]{.45\textwidth}
    \begin{center}  
  \includegraphics[width=\columnwidth]{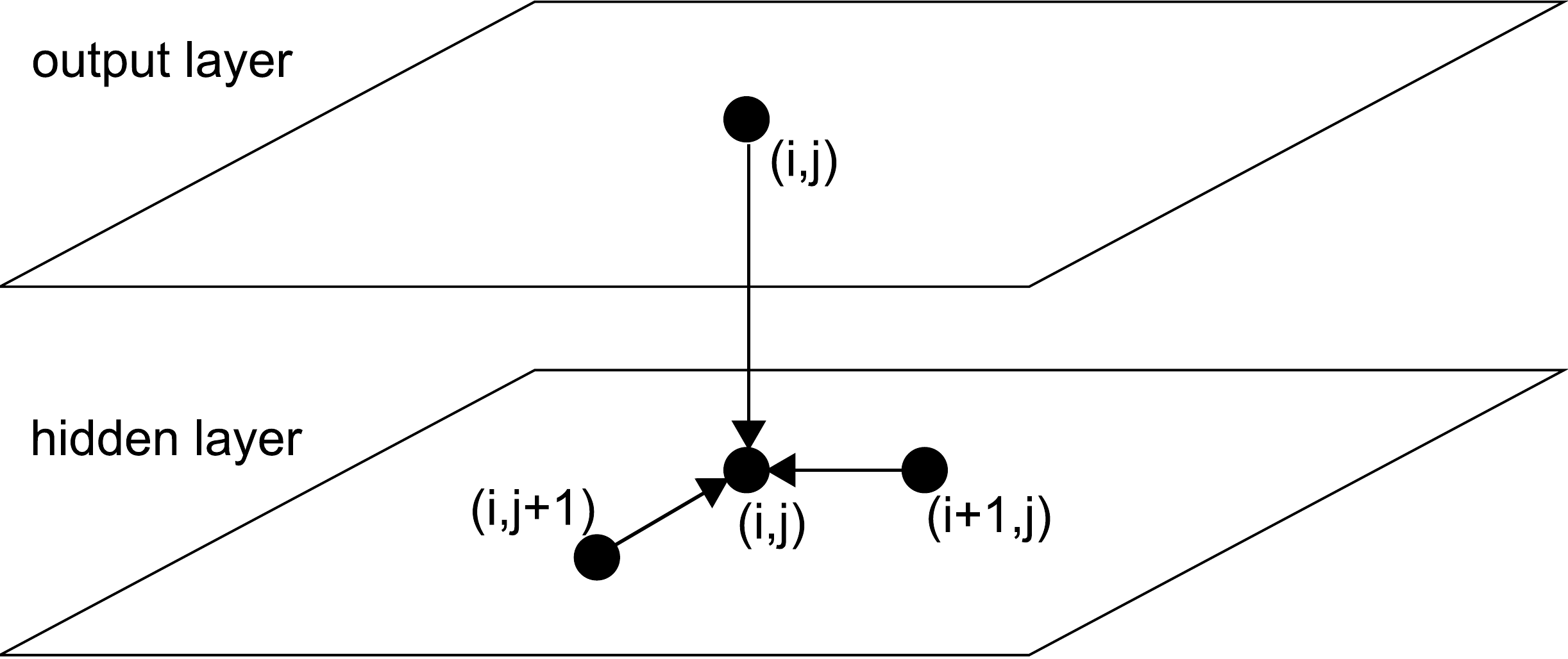}
	\caption{2D RNN Backward pass.} 
	\label{fig:mdrnn_backward}
    \end{center}
  \end{minipage}
\end{figure}

The basic idea of MDRNNs is to replace the single recurrent connection found in standard RNNs with as many recurrent connections as there are dimensions in the data. During the forward pass, at each point in the data sequence, the hidden layer of the network receives both an external input and its own activations from one step back along all dimensions. Figure \ref{fig:mdrnn_forward} illustrates the two dimensional case.

Note that, although the word \emph{sequence} usually connotes one dimensional data, we will use it to refer to data examplars of any dimensionality. For example, an image is a two dimensional sequence, a video is a three dimensional sequence, and a series of fMRI brain scans is a four dimensional sequence.


\begin{figure}
\begin{center}
  \includegraphics[width=0.4\columnwidth]{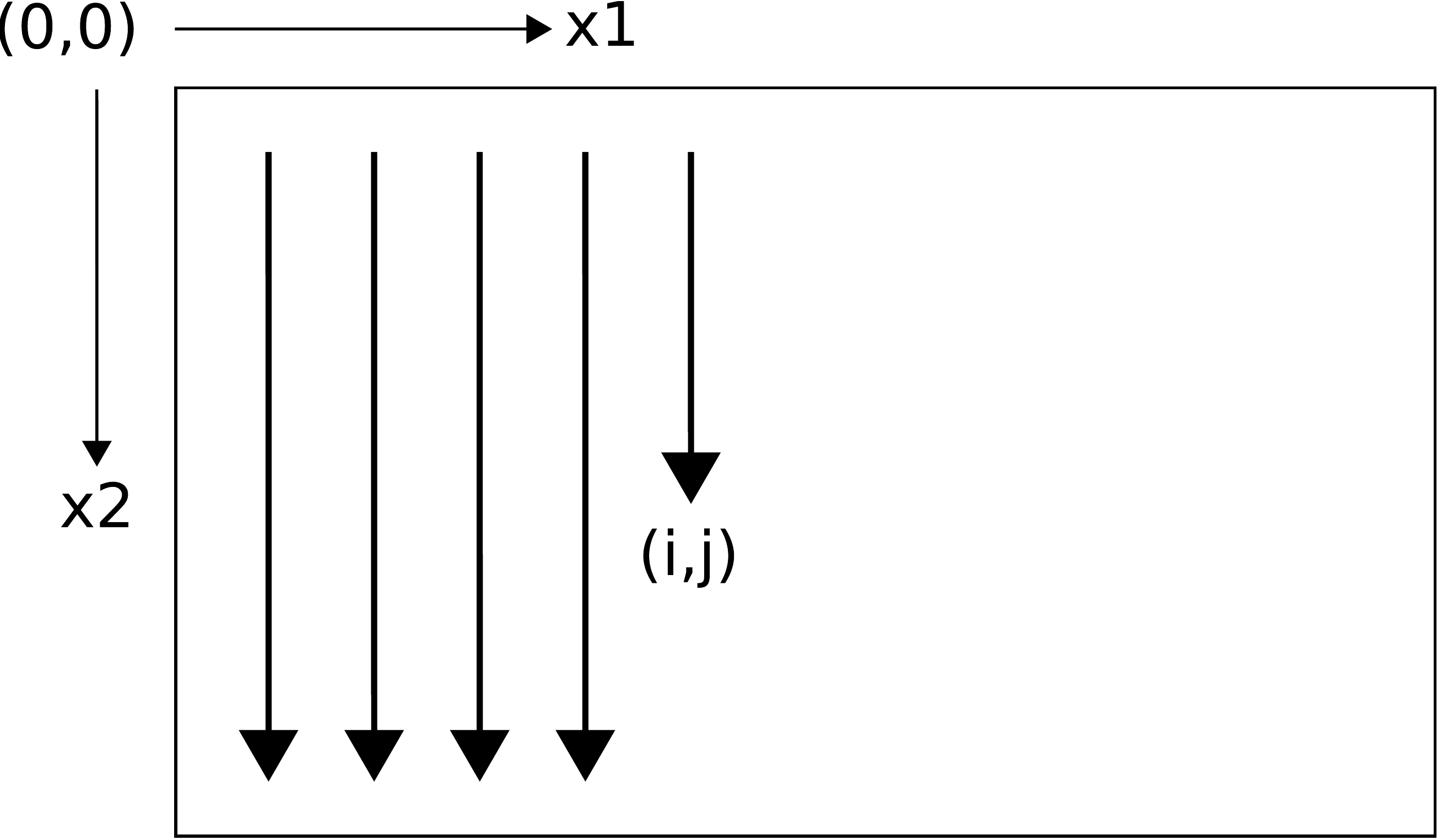}
	\caption{2D sequence ordering. The MDRNN forward pass starts at the origin and follows the direction of the arrows. The point (i,j) is never reached before both (i-1,j) and (i,j-1).}
	\label{fig:mdrnn_ordering}
\end{center}
\end{figure}

Clearly, the data must be processed in such a way that when the network reaches a point in an n-dimensional sequence, it has already passed through all the points from which it will receive its previous activations. This can be ensured by following a suitable ordering on the points $\{(x_1,x_2,...,x_n)\}$. One example of a suitable ordering is $(x_1,\dots,x_n) < (x'_1,\dots,x'_n)$ if $\exists\ m \in (1,\dots,n)$ such that $x_m < x'_m$ and $x_i = x'_i \ \forall\ i \in (1,\dots,m-1)$. Note that this is not the only possible ordering, and that its realisation for a particular sequence depends on an arbitrary choice of axes. We will return to this point in Section \ref{sec:bidirectional}. Figure \ref{fig:mdrnn_ordering} illustrates the ordering for a 2 dimensional sequence.

The forward pass of an MDRNN can then be carried out by feeding forward the input and the $n$ previous hidden layer activations at each point in the ordered input sequence, and storing the resulting hidden layer activations. Care must be taken at the sequence boundaries not to feed forward activations from points outside the sequence.

Note that each `point' in the input sequence will in general be a multivalued vector. For example, in a two dimensional colour image, the inputs could be single pixels represented by RGB triples, or blocks of pixels, or the outputs of a preprocessing method such as a discrete cosine transform.

The error gradient of an MDRNN (that is, the derivative of some objective function with respect to the network weights) can be calculated with an n-dimensional extension of the backpropagation through time (BPTT \cite{williams+zipser:1995}) algorithm. As with one dimensional BPTT, the sequence is processed in the reverse order of the forward pass. At each timestep, the hidden layer receives both the output error derivatives and its own $n$ `future' derivatives. Figure \ref{fig:mdrnn_backward} illustrates the BPTT backward pass for two dimensions. Again, care must be taken at the sequence boundaries.


At a point $\mathbf{x} = (x_1,\dots,x_n)$ in an n-dimensional sequence, define $i_j^{\mathbf{x}}$ and $h_k^{\mathbf{x}}$ respectively as the activations of the $j^{th}$ input unit and the $k^{th}$ hidden unit. Define $w_{kj}$ as the weight of the connection going from unit $j$ to unit $k$. Then for an n-dimensional MDRNN whose hidden layer consists of summation units with the $tanh$ activation function, the forward pass for a sequence with dimensions $(X_1,X_2,\dots,X_n)$ can be summarised as follows:

\begin{algorithm}
\caption{MDRNN Forward Pass}
\begin{algorithmic}
\FOR{$x_1=0$ to $X_1-1$}
\FOR{$x_2=0$ to $X_2-1$}
\STATE \dots
\FOR{$x_n=0$ to $X_n-1$}
\STATE initialize $a \leftarrow \sum_{j}{{in}^{\mathbf{x}}_j w_{kj}}$
\FOR{$i = 1$ to $n$}
\IF{$x_i > 0$}
\STATE  $a \leftarrow a + \sum_{j}{h^{(x_1,\dots,x_i-1,\dots,x_n)}_j w_{kj}}$
\ENDIF
\ENDFOR
\STATE $h_k^{\mathbf{x}} \leftarrow tanh(a)$
\ENDFOR

\ENDFOR
\ENDFOR
\end{algorithmic}
\label{alg:mdrnn_forward}
\end{algorithm}

Defining $\hat{o}_j^{\mathbf{x}}$ and $\hat{h}_k^{\mathbf{x}}$ respectively as the derivatives of the objective function with respect to the  activations of the $j^{th}$ output unit and the $k^{th}$ hidden unit at point $\mathbf{x}$, the backward pass is:

\begin{algorithm}
\caption{MDRNN Backward Pass}
\begin{algorithmic}
\FOR{$x_1=X_1-1$ to $0$}
\FOR{$x_2=X_2-1$ to $0$}
\STATE \dots
\FOR{$x_n=X_n-1$ to $0$}
\STATE initialize $e \leftarrow \sum_{j}{\hat{o}^{\mathbf{x}}_j w_{jk}}$
\FOR{$i = 1$ to $n$}
\IF{$x_i < X_i - 1$}
\STATE  $e \leftarrow e + \sum_{j}{\hat{h}^{(x_1,\dots,x_i+1,\dots,x_n)}_j w_{jk}}$
\ENDIF
\ENDFOR
\STATE $\hat{h}_k^{\mathbf{x}} \leftarrow tanh'(e)$
\ENDFOR

\ENDFOR
\ENDFOR
\end{algorithmic}
\label{alg:mdrnn_backward}
\end{algorithm}

Since the forward and backward pass require one pass each through the data sequence, the overall complexity of MDRNN training is linear in the number of data points and the number of network weights.

\subsection{Multi-directional MDRNNs}\label{sec:bidirectional}

\begin{figure}
  \begin{minipage}[t]{.45\textwidth}
    \begin{center}  
  \includegraphics[width=0.8\columnwidth]{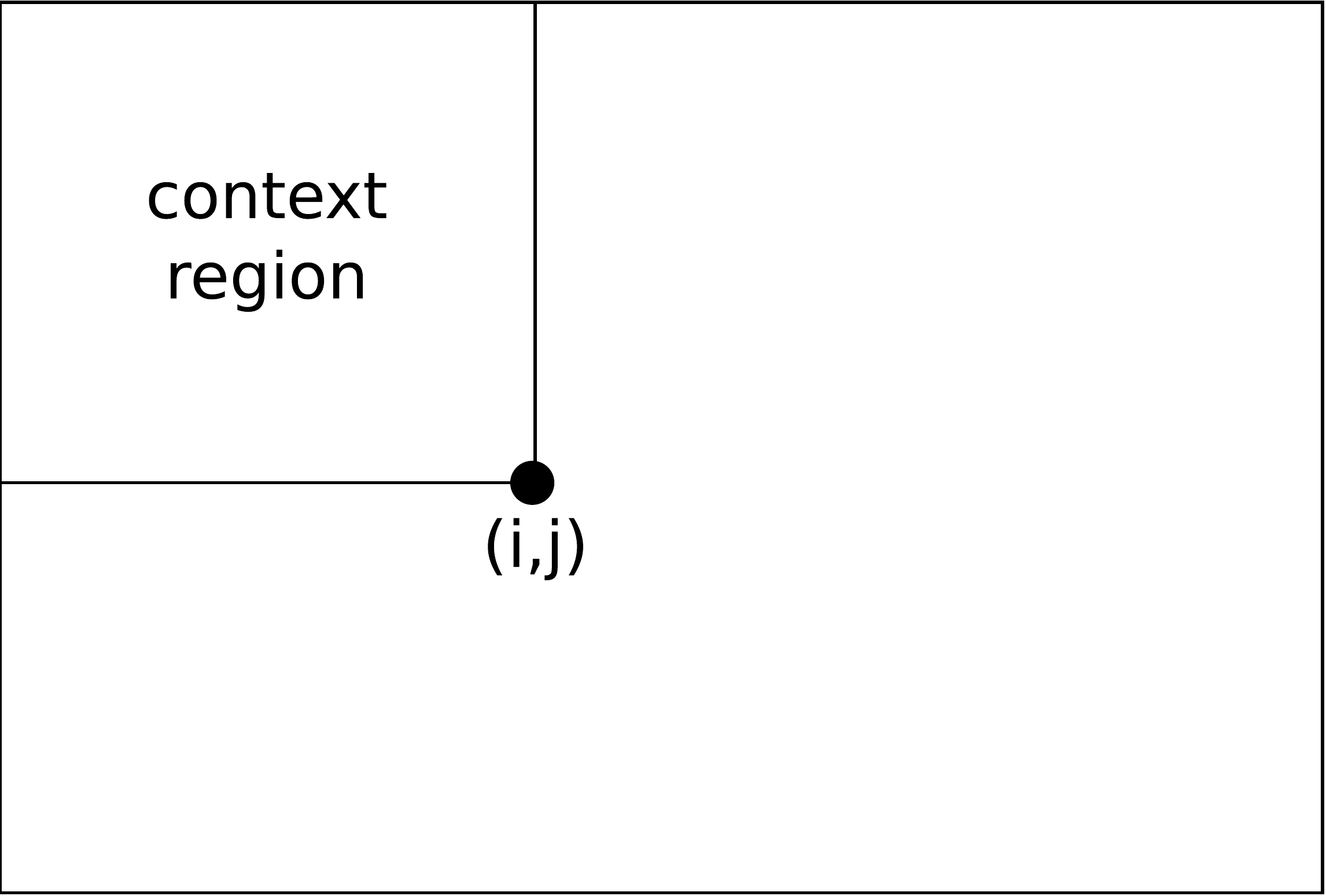}
	\caption{Context available at (i,j) to a 2D RNN with a single hidden layer.}
	\label{fig:md_uni_context}
    \end{center}
  \end{minipage}
  \hfill
  \begin{minipage}[t]{.45\textwidth}
    \begin{center}  
      \includegraphics[width=0.8\columnwidth]{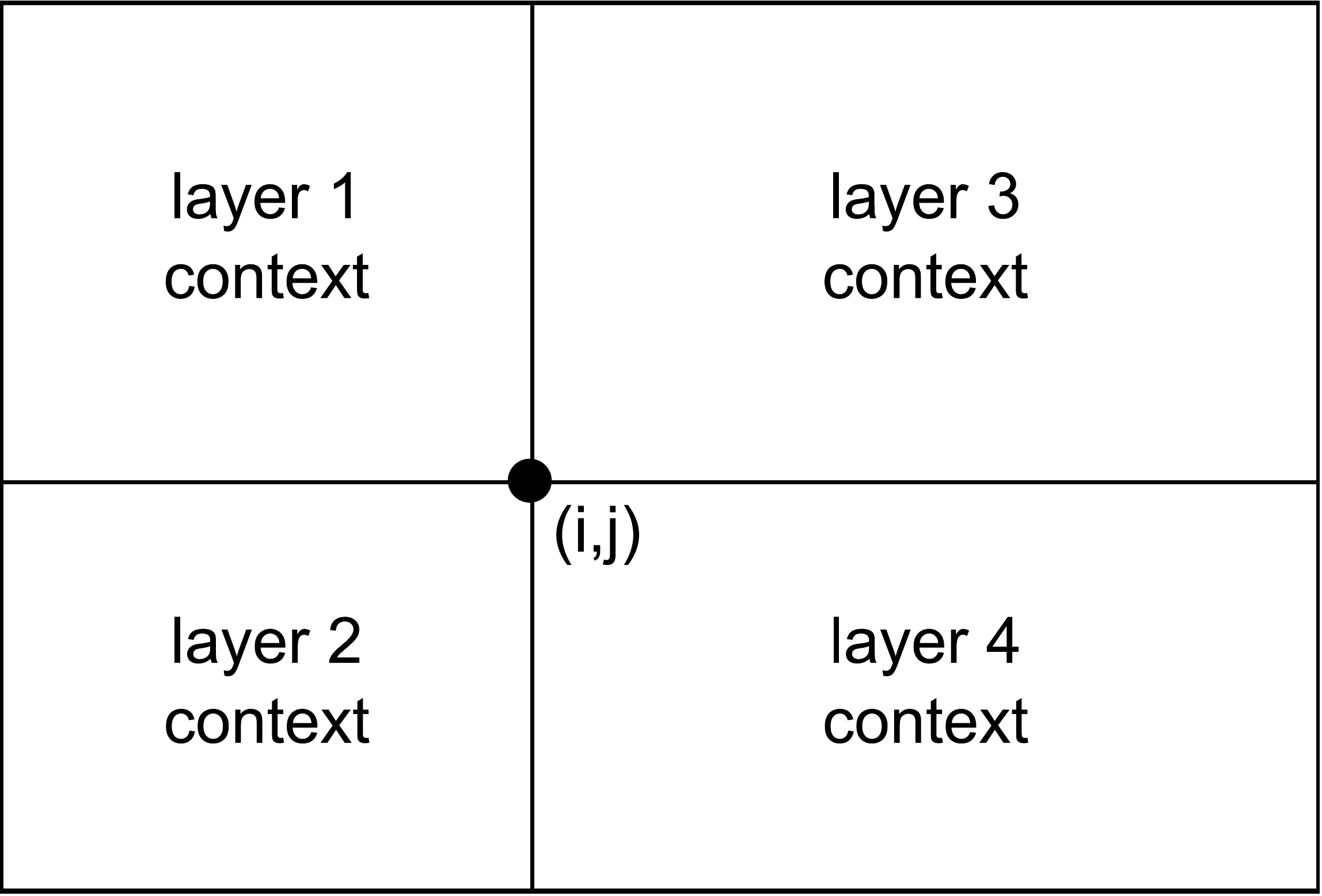}
      \caption{Context available at (i,j) to a multi-directional 2D RNN.} 
      \label{fig:md_multi_context}
    \end{center}
  \end{minipage}
\end{figure}

At a point $(x_1,...,x_n)$ in the input sequence, the network described above has access to all points $(x'_1,...,x'_n)$ such that $x'_i \leq x_i \forall\ i \in (1,...,n)$. This defines an n-dimensional `context region' of the full sequence, as shown in Figure \ref{fig:md_uni_context}. For some tasks, such as object recognition, this would in principal be sufficient. The network could process the image as usual, and output the object label at a point when the object to be recognized is entirely contained in the context region.


Intuitively however, we would prefer the network to have access to the surrounding context in all directions. This is particularly true for tasks where precise localization is required, such as image segmentation.

For one dimensional RNNs, the problem of multi-directional context was solved in 1997 by the introduction of bidirectional recurrent neural networks (BRNNs) \cite{schuster97bidirectional}. BRNNs contain two separate hidden layers that process the input sequence in the forward and reverse directions. The two hidden layers are connected to a single output layer, 
thereby providing the network with access to both past and future context.

\begin{figure}
\begin{center}
  \includegraphics[width=0.4\columnwidth]{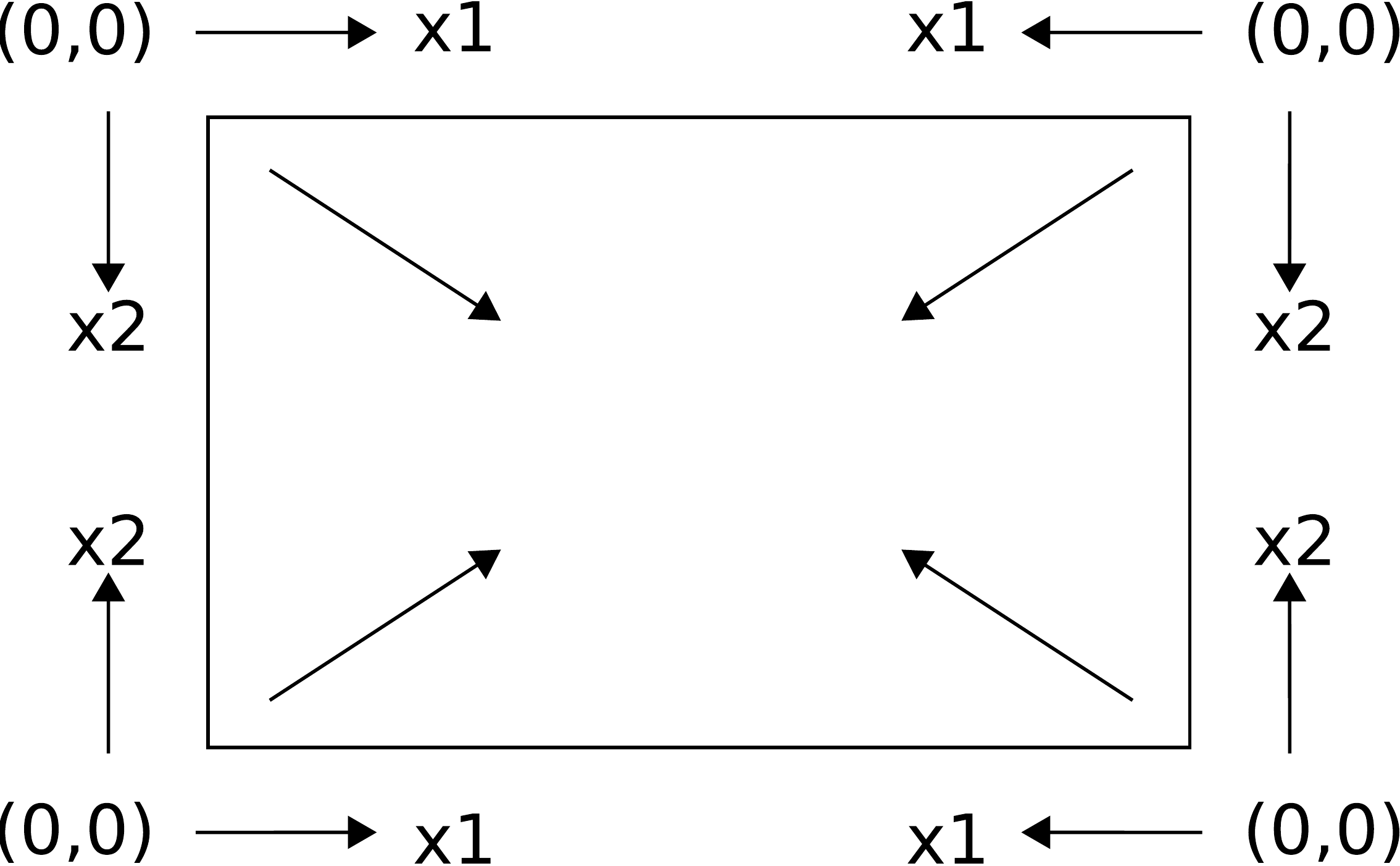}
	\caption{Axes used by the 4 hidden layers in a multi-directional 2D network.	The arrows inside the rectangle indicate the direction of propagation during the forward pass.}
	\label{fig:md_axes}
\end{center}
\end{figure}

BRNNs can be extended to n-dimensional data by using $2^n$ separate hidden layers, each of which processes the sequence using the ordering defined above, but with a different choice of axes. More specifically, the axes are chosen so that their origins lie on the $2^n$ vertices of the sequence. The 2 dimensional case is illustrated in Figure \ref{fig:md_axes}. As before, the hidden layers are connected to a single output layer, which now has access to all surrounding context (see Figure \ref{fig:md_multi_context}).

If the size of the hidden layers is held constant, multi-directional MDRNNs scales as $O(2^n)$ for n-dimensional data. In practice however, we have found that using $2^n$ small layers gives better results than 1 large layer with the same overall number of weights, presumably because the data processing is shared between the hidden layers. This also holds in one dimension, as previous experiments have demonstrated \cite{graves05nn}. In any case the complexity of the algorithm remains linear in the number of data points and the number of parameters, and the number of parameters is independent of the data dimensionality.

For a multi-directional MDRNN, the forward and backward passes through an n-dimensional sequence can be summarised as follows:
\begin{algorithm}[H]
\caption{Multi-directional MDRNN Forward Pass}
\begin{algorithmic}[1]
\STATE For each of the $2^n$ hidden layers choose a distinct vertex of the sequence, then define a set of axes such that the vertex is the origin and all sequence co-ordinates are $\geq 0$
\STATE Repeat Algorithm \ref{alg:mdrnn_forward} for each hidden layer
\STATE At each point in the sequence, feed forward all
hidden layers to the output layer
\end{algorithmic}
\end{algorithm}
\begin{algorithm}[H]
\caption{Multi-directional MDRNN Backward Pass}
\begin{algorithmic}[1]
\STATE At each point in the sequence, calculate the derivative of the objective function with respect to the activations of output layer
\STATE With the same axes as above, repeat Algorithm \ref{alg:mdrnn_backward} for each hidden layer
\end{algorithmic}
\end{algorithm}
\subsection{Multi-dimensional Long Short-Term Memory}\label{sec:lstm}

So far we have implicitly assumed that the network can make use of all context to which it has access. For standard RNN architectures however, the range of context that can practically be used is limited. The problem is that the influence of a given input on the hidden layer, and therefore on the network output, either decays or blows up exponentially as it cycles around the network's recurrent connections. This is usually referred to as the \emph{vanishing gradient problem} \cite{Hochreiter:01book}.

Long Short-Term Memory (LSTM)~\cite{hochreiter+schmidhuber:1997,gers+schraudholph+schmidhuber:2002} is an RNN architecture specifically designed to address the vanishing gradient problem. An LSTM hidden layer consists of multiple recurrently connected subnets, known as memory blocks. Each block contains a set of internal units, known as cells, whose activation is controlled by three multiplicative units: the input gate, forget gate and output gate. The effect of the gates is to allow the cells to store and access information over long periods of time, thereby avoiding the vanishing gradient problem. 


The standard formulation of LSTM is explicitly one-dimensional, since the cell contains a single self connection, whose activation is controlled by a single forget gate. However we can easily extend this to $n$ dimensions by using instead $n$ self connections (one for each of the cell's previous states along every dimension) with $n$ forget gates.

\section{Experiments}\label{sec:exps}



\subsection{Air Freight Data}

\begin{figure}
\begin{center}
\includegraphics[width=0.25\columnwidth]{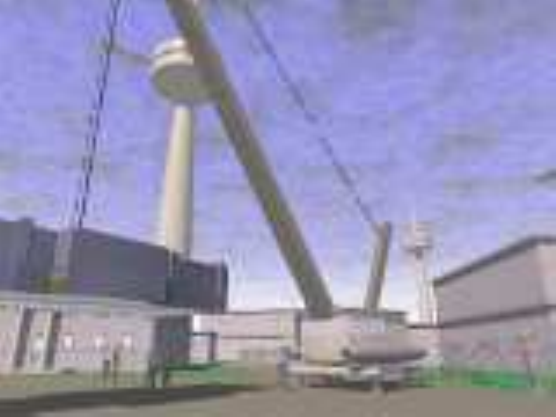}
\hspace{.2in}
\includegraphics[width=0.25\columnwidth]{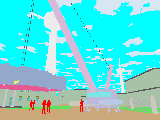}
\end{center}
\caption{Frame from the Air Freight database, showing the original image (left) and the colour-coded texture segmentation (right).}
\label{fig:air_freight}
\end{figure}

The Air Freight database is a ray-traced colour image sequence that comes with a ground truth segmentation based on textural characteristics (Figure \ref{fig:air_freight}). The sequence is 455 frames long and contains 155 distinct textures. Each frame is 120 pixels high and 160 pixels wide.

The advantage of ray-traced data is the true segmentation can be defined directly from the 3D models. Although the images are not real, they are realistic in the sense that they have significant lighting, specular effects etc.

We used the sequence to define a 2D image segmentation task, where the aim was to assign each pixel in the input data to the correct texture class. We divided the data at random into a 250 frame train set, a 150 frame test set and a 55 frame validation set. Note that we could have instead defined a 3D task where the network processed the entire video as one sequence. However, this would have left us with only one exemplar.

For this task we used a multi-directional MDRNN with 4 LSTM hidden layers. Each layer consisted of 25 memory blocks, each containing 1 cell, 2 forget gates, 1 input gate, 1 output gate and 5 peephole weights. This gave a total 600 hidden units. The input and output activation functions of the cells were both tanh, and the activation function for the gates was the logistic sigmoid in the range $[0,1]$. The input layer was size 3 (one each for the red, green and blue components of the pixels) and the output layer was size 155 (one unit for each textural class). The network contained 43,257 trainable weights in total. As is standard for classification tasks, the softmax activation function was used at the output layer, with the cross-entropy objective function \cite{bridleSoftmax1990}. The network was trained using online gradient descent (weight updates after every training sequence) with a learning rate of $10^{-6}$ and a momentum of $0.9$.

The final pixel classification error rate was $7.3\,\%$ on the test set.

\subsection{MNIST Data}

The MNIST database \cite{lecun-98} of isolated handwritten digits is a subset of a larger database available from NIST. It consists of size-normalized, centered images, each of which is 28 pixels high and 28 pixels wide and contains a single handwritten digit. The data comes divided into a training set with 60,000 images and a test set with 10,000 images. We used 10,000 of the training images for validation, leaving 50,000 for training.


The usual task on MNIST is to label the images with the corresponding digits. This is benchmark task for which many algorithms have been evaluated.

We carried out a slightly modified task where each pixel was classified according to the digit it belonged to, with an additional class for background pixels. However, the original task can be recovered by simply choosing the digit whose corresponding output unit had the highest cumulative activation for the entire sequence.

\begin{figure}[t]
\begin{center}
\includegraphics[width=0.2\columnwidth]{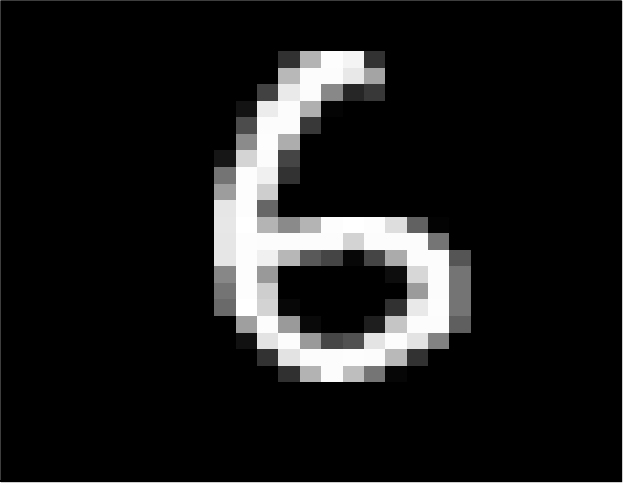}
\hspace{.2in}
\includegraphics[width=0.2\columnwidth]{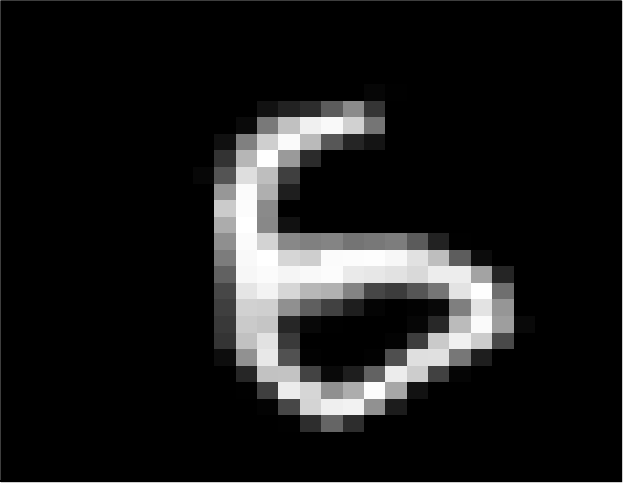}
\end{center}
\caption{MNIST image before and after deformation.}
\label{fig:mnist_deformed}
\end{figure}

To test the network's robustness to input warping, we also evaluated it on an altered version of the MNIST test set, where elastic deformations had been applied to every image (Figure \ref{fig:mnist_deformed}).

We compared our results with the convolution neural network that has achieved the best results so far on MNIST \cite{simard03icdar}. Note that we re-implemented the convolution network ourselves, and we did not augment the training set with elastic distortions, which gives a substantial improvement in performance.

The MDRNN for this task was identical to that for the Air Freight task with the following exceptions: the sizes of the input and output layers were now 1 (for grayscale pixels) and 11 (one for each digit, plus background) respectively, giving 27,511 weights in total, and the gradient descent learning rate was $10^{-5}$.

For the distorted test set, we used the same degree of elastic deformation used by Simard \cite{simard03icdar} to augment the training set ($\sigma=4.0$, $\alpha=34.0$), with a different initial random field for every sample image. 

Table \ref{tab:mnist_results} shows thatlthough the MDRNN performed slightly worse on the clean test set, its performance was considerably better on the warped test set. This suggests that MDRNNs are more robust to input warping than convolution networks.

\begin{table}
\begin{center}
\caption{Image error rates on MNIST (pixel error rates in brackets)}
\begin{tabular}{lll}
\hline
\noalign{\smallskip}
Algorithm & Clean Test Set & Warped Test Set \\
\noalign{\smallskip}
\hline
\noalign{\smallskip}
MDRNN & 1.1\,\% (0.5\,\%) & 6.8\,\% (3.8\,\%)\\
Convolution & 0.9\,\% & 11.3\,\%\\
\hline
\end{tabular}
\label{tab:mnist_results}
\end{center}
\end{table}


\subsection{Analysis}

\begin{figure}[t]
\begin{center}
\includegraphics[width=\columnwidth]{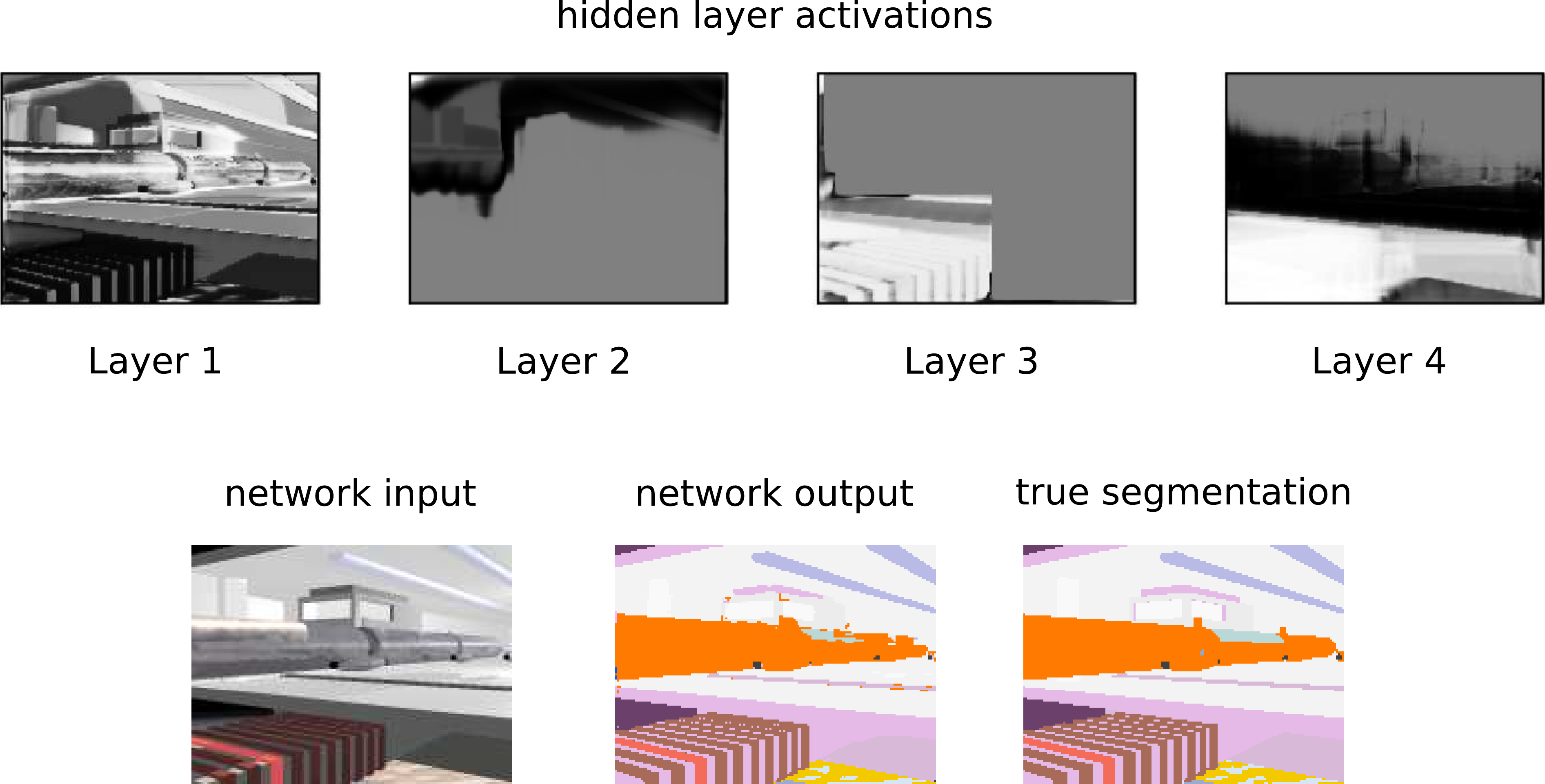}
\end{center}
\caption{2D MDRNN applied to an image from the Air Freight database. The hidden layer activations display one unit from each of the layers. A common behaviour is to `mask off' parts of the image, exhibited here by layers 2 and 3.}
\label{fig:air_freight_acts}
\end{figure}

One benefit of two dimensional tasks is that the operation of the network can be easily visualised. Figure \ref{fig:air_freight_acts} shows the network activations during a frames from the Air Freight database. As can be seen, the network segments this image almost perfectly, in spite of difficult, reflective surfaces such as the glass and metal tube running from left to right. Clearly, classifying individual pixels in such a surface requires considerable use of context.

\begin{figure}[t]
\begin{center}
\includegraphics[width=0.9\columnwidth]{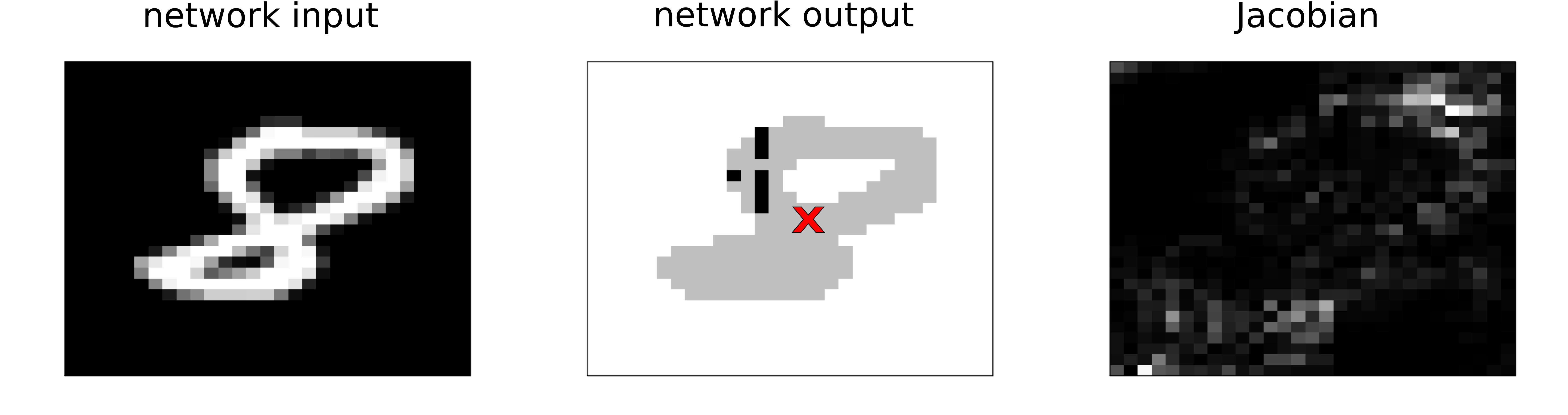}
\end{center}
\caption{Jacobian matrix of a 2D RNN for an image from the MNIST database. 
The white outputs correspond to the class `background' and the light grey ones to `8'. The black outputs represent misclassifications. The output pixel for which the Jacobian is calculated is marked with a cross. Absolute values are plotted for the Jacobian, and lighter colours are used for higher values.}
\label{fig:mnist_jacobian}
\end{figure}

A precise measure of the network's sensitivity to context can be found by analysing the derivatives of the network outputs at a particular point $\mathbf{x}$ in the sequence with respect to the inputs at all points $\mathbf{x}'$ in the sequence. The matrix $\frac{\partial o^{\mathbf{x}}_k}{\partial in^{\mathbf{x}'}_j}$ of these derivatives is referred to as the \emph{Jacobian} matrix. Figure \ref{fig:mnist_jacobian} shows the absolute value of the Jacobian matrix for a single output during classification of an image from the MNIST database. It can be seen that the network responds to context from across the entire image, and seems particularly attuned to the outline of the digit.

\section{Conclusion}\label{sec:conclusion}

We have introduced multi-dimensional recurrent neural networks (MDRNNs), thereby extending the applicabilty of RNNs to n-dimensional data. We have added multi-directional hidden layers that provide the network with access to all contextual information, and we have developed a multi-dimensional variant of the Long Short-Term Memory RNN architecture. We have tested MDRNNs on two image segmentation tasks, and found that it was more robust to input warping than a state-of-the-art digit recognition algorithm.

\bibliographystyle{plain}
{\bibliography{alex}}

\end{document}